\documentclass[11pt,a4paper]{article}
\usepackage[nohyperref]{acl2019}
\usepackage{times}
\usepackage{latexsym}
\usepackage{graphicx}
\usepackage{graphics, amsmath}
\usepackage{amsfonts}
\usepackage{amsthm}
\usepackage{amssymb}
\usepackage{multirow}
\usepackage{enumitem, natbib}
\usepackage{url}
\usepackage{makecell}

\aclfinalcopy 
\def\aclpaperid{848} 

\title{Multi-hop Reading Comprehension across Multiple Documents by Reasoning over Heterogeneous Graphs}

\author{Ming Tu, Guangtao Wang, Jing Huang, Yun Tang, Xiaodong He, Bowen Zhou \\
  JD AI Research \\
  \texttt{\{ming.tu,guangtao.wang,jing.huang,yun.tang,xiaodong.he,bowen.zhou\}@jd.com} \\}

\date{}

\begin{document}
\maketitle
\begin{abstract}
  Multi-hop reading comprehension (RC) across documents poses new challenge over single-document RC because it requires reasoning over multiple documents to reach the final answer. In this paper, we propose a new model to tackle the multi-hop RC problem. We introduce a heterogeneous graph with different types of nodes and edges, which is named as Heterogeneous Document-Entity (HDE) graph. The advantage of HDE graph is that it contains different granularity levels of information including candidates, documents and entities in specific document contexts. Our proposed model can do reasoning over the HDE graph with nodes representation initialized with co-attention and self-attention based context encoders. We employ Graph Neural Networks (GNN) based message passing algorithms to accumulate evidences on the proposed HDE graph. Evaluated on the blind test set of the Qangaroo
  \textsc{WikiHop} data set, our HDE graph based single model delivers competitive result, and the ensemble model achieves the state-of-the-art performance.
\end{abstract}

\section{Introduction}
Being able to comprehend a document and output correct answer given a query/question about content in the document, often referred as machine reading comprehension (RC) or question answering (QA), is an important and challenging task in natural language processing (NLP). Plenty of data sets have been constructed to facilitate research on this topic, such as SQuAD \cite{rajpurkar2016squad, rajpurkar2018know}, NarrativeQA \cite{kovcisky2018narrativeqa} and CoQA \cite{reddy2018coqa}. Many neural models have been proposed to tackle the machine RC/QA problem \cite{seo2016bidirectional, xiong2016dynamic, tay2018densely}, and great success has been achieved, especially after the release of the BERT \cite{devlin2018bert}.

\begin{figure}[t]
\framebox{
\parbox{0.45\textwidth}{
\small
\textbf{Query:} record\_label \textcolor{orange}{get ready} \newline
\textbf{Support doc 1:} Mason Durell Betha (born August 27, 1977), better known by stage name \textcolor{purple}{Mase} (formerly often stylized Ma\$e or MA\$E), is an American hip hop recording artist and minister. He is best known for being signed to Sean ``Diddy'' Combs's label \textcolor{red}{Bad Boy Records}. \dots \smallskip\newline
\textbf{Support doc 2:} \textcolor{orange}{``Get Ready''} was the only single released from \textcolor{purple}{Mase's} second album, Double Up. It was released on May 25, 1999, produced by \textcolor{blue}{Sean ``Puffy'' Combs}, Teddy Riley and Andreao ``Fanatic'' Heard and featured R\&B group, Blackstreet, it contains a sample of ``A Night to Remember'', performed by Shalamar. \dots \smallskip\newline
\textbf{Support doc 3:} Bad Boy Entertainment (also known as \textcolor{red}{Bad Boy Records}) is an American record label founded in 1993 by \textcolor{blue}{Sean Combs}. \dots \smallskip\newline
\textbf{Candidates:} \textcolor{red}{bad boy records}, record label, rock music, \dots \newline
\textbf{Answer:} \textcolor{red}{bad boy records}
}
}
\caption{A \textsc{WikiHop} example. Words with different colors indicate the evidences across documents.} 
\label{fig:wikihop}
\end{figure}

However, current research mainly focuses on machine RC/QA on a single document or paragraph, and still lacks the ability to do reasoning across multiple documents when a single document is not enough to find the correct answer. To promote the study for multi-hop RC over multiple documents, two data sets are recently proposed: \textsc{WikiHop} \cite{welbl2018constructing} and HotpotQA \cite{yang2018hotpotqa}. These two data sets require multi-hop reasoning over multiple supporting documents to find the answer. In Figure~\ref{fig:wikihop}, we show an excerpt from one sample in \textsc{WikiHop} development set to illustrate the need for multi-hop reasoning. 


Two types of approaches have been proposed on the multi-hop multi-document RC problem. The first is based on previous neural RC models. The earliest attempt in \cite{dhingra2018neural} concatenated all supporting documents and designed a recurrent layer to explicitly exploit the skip connections between entities given automatically generated coreference annotations. Adding this layer to the neural RC models improved performance on multi-hop tasks. Recently, an attention based system \cite{zhong2019coarse} utilizing both document-level and entity-level information achieved state-of-the-art results on \textsc{WikiHop} data set, proving that techniques like co-attention and self-attention widely employed in single-document RC tasks are also useful in multi-document RC tasks.

The second type of research work is based on graph neural networks (GNN) for multi-hop reasoning. The study in \citet{song2018exploring} adopted two separate name entity recognition (NER) and coreference resolution systems to locate entities in support documents. Those entities serve as nodes in GNN to enable multi-hop reasoning across documents. Work in \citet{de2018question} directly used mentions of candidates (found in documents by simple exact matching strategy) as GNN nodes and calculate classification scores over mentions of candidates.

In this paper, we propose a new method to solve the multi-hop RC problem across multiple documents. Inspired by the success of GNN based methods \cite{song2018exploring, de2018question} for multi-hop RC, we introduce a new type of graph, called Heterogeneous Document-Entity (HDE) graph. Our proposed HDE graph has the following advantages:
\begin{itemize}
    \item Instead of graphs with single type of nodes \cite{song2018exploring,de2018question}, the HDE graph contains different types of query-aware nodes representing different granularity levels of information. Specifically, instead of only entity nodes as in \cite{song2018exploring, de2018question}, we include nodes corresponding to candidates, documents and entities. In addition, following the success of Coarse-grain Fine-grain Coattention (CFC) network \cite{zhong2019coarse}, 
    we apply both co-attention and self-attention to learn query-aware node representations of candidates, documents and entities; 
    \item The HDE graph enables rich information interaction among different types of nodes thus facilitate accurate reasoning. Different types of nodes are connected with different types of edges to highlight the various structural information presented among query, document and candidates. 
\end{itemize}
Through ablation studies, we show the effectiveness of our proposed HDE graph for multi-hop multi-document RC task. Evaluated on the blind test set of \textsc{WikiHop}, our proposed end-to-end trained {\em single} neural model beats the current published state-of-the-art results in \cite{zhong2019coarse} and is the 2nd best model on the \textsc{WikiHop} leaderboard. Meanwhile, our {\em ensemble} model ranks 1st place on the \textsc{WikiHop} leadrboard and surpasses the human performance (as reported in \cite{welbl2018constructing}) on this data set by 0.2\% \footnote{By May 30th 2019, \url{http://qangaroo.cs.ucl.ac.uk/leaderboard.html}}. This is achieved without using pretrained contextual ELMo embedding \cite{peters2018deep}.

\section{Related Work}
The study presented in this paper is directly related to existing research on multi-hop reading comprehension across multiple documents \cite{dhingra2018neural, song2018exploring, de2018question, zhong2019coarse, kundu2018exploiting}. The method presented in this paper is similar to previous studies using GNN for multi-hop reasoning \cite{song2018exploring, de2018question}. Our novelty is that we propose to use a heterogeneous graph instead of a graph with single type of nodes to incorporate different granularity levels of information. The co-attention and self-attention based encoding of multi-level information presented in each input is also inspired by the CFC model \cite{zhong2019coarse} because they show the effectiveness of attention mechanisms. Our model is very different from the other two studies \cite{dhingra2018neural,kundu2018exploiting}: these two studies both explicitly score the possible reasoning paths with extra NER or coreference resolution systems while our method does not require these modules and we do multi-hop reasoning over graphs. Besides these studies, our work is also related to the following research directions.

\textbf{Multi-hop RC:} There exist several different data sets that require reasoning in multiple steps in literature, for example bAbI \cite{weston2015towards}, MultiRC \cite{MultiRC2018} and OpenBookQA \cite{mihaylov2018can}. A lot of systems have been proposed to solve the multi-hop RC problem with these data sets \cite{sun2018improving, wu2019global}. However, these data sets require multi-hop reasoning over multiple sentences or multiple common knowledge while the problem we want to solve in this paper requires collecting evidences across multiple documents.

\textbf{GNN for NLP:} Recently, there is considerable amount of interest in applying GNN to NLP tasks and great success has been achieved. For example, in neural machine translation, GNN has been employed to integrate syntactic and semantic information into encoders \cite{bastings2017graph, marcheggiani2018exploiting}; \citet{zhang2018graph} applied GNN to relation extraction over pruned dependency trees; the study by \citet{yao2018graph} employed GNN over a heterogeneous graph to do text classification, which inspires our idea of the HDE graph; \citet{liu2018contextualized} proposed a new contextualized neural network for sequence learning by leveraging various types of non-local contextual information in the form of information passing over GNN. These studies are related to our work in the sense that we both use GNN to improve the information interaction over long context or across documents.

\section{Methodology}

In this section, we describe different modules of the proposed Heterogeneous Document-Entity (HDE) graph-based multi-hop RC model. The overall system diagram is shown in Figure~\ref{fig:1}. Our model can be roughly categorized into three parts: initializing HDE graph nodes with co-attention and self-attention based context encoding, reasoning over HDE graph with GNN based message passing algorithms and score accumulation from updated HDE graph nodes representations.

\subsection{Context encoding}

Given a query $q$ with the form of (\textit{s}, \textit{r}, \textit{?}) which represents subject, relation and unknown object respectively, a set of support documents $S_q$ and a set of candidates $C_q$, the task is to predict the correct answer $a^{*}$ to the query. To encode information including in the text of query, candidates and support documents, we use a pretrained embedding matrix \cite{pennington2014glove} to convert word sequences to sequences of vectors. Let $\mathbf{X}_q \in \mathbb{R}^{l_q \times d}$, $\mathbf{X}_s^i \in \mathbb{R}^{l_s^i \times d}$ and $\mathbf{X}_c^j \in \mathbb{R}^{l_c^j \times d}$ represent the embedding matrices of query, $i$-th supporting document and $j$-th candidate of a sample, where $l_q$, $l_s^i$ and $l_c^j$ are the numbers of words in query, $i$-th supporting document and $j$-th candidate respectively. $d$ is the dimension of the word embedding. We use bidirectional recurrent neural networks (RNN) with gated recurrent unit (GRU) \cite{cho2014learning} to encode the contextual information present in the query, supporting documents and candidates separately. The output of query, document and candidate encoders are $\mathbf{H}_q \in \mathbb{R}^{l_q \times h}$, $\mathbf{H}_s^i \in \mathbb{R}^{l_s^i \times h}$ and $\mathbf{H}_c^j \in \mathbb{R}^{l_c^j \times h}$. $h$ denotes the output dimension of RNN encoders.

\begin{figure}[t]
    \centering
    \includegraphics[width=\linewidth]{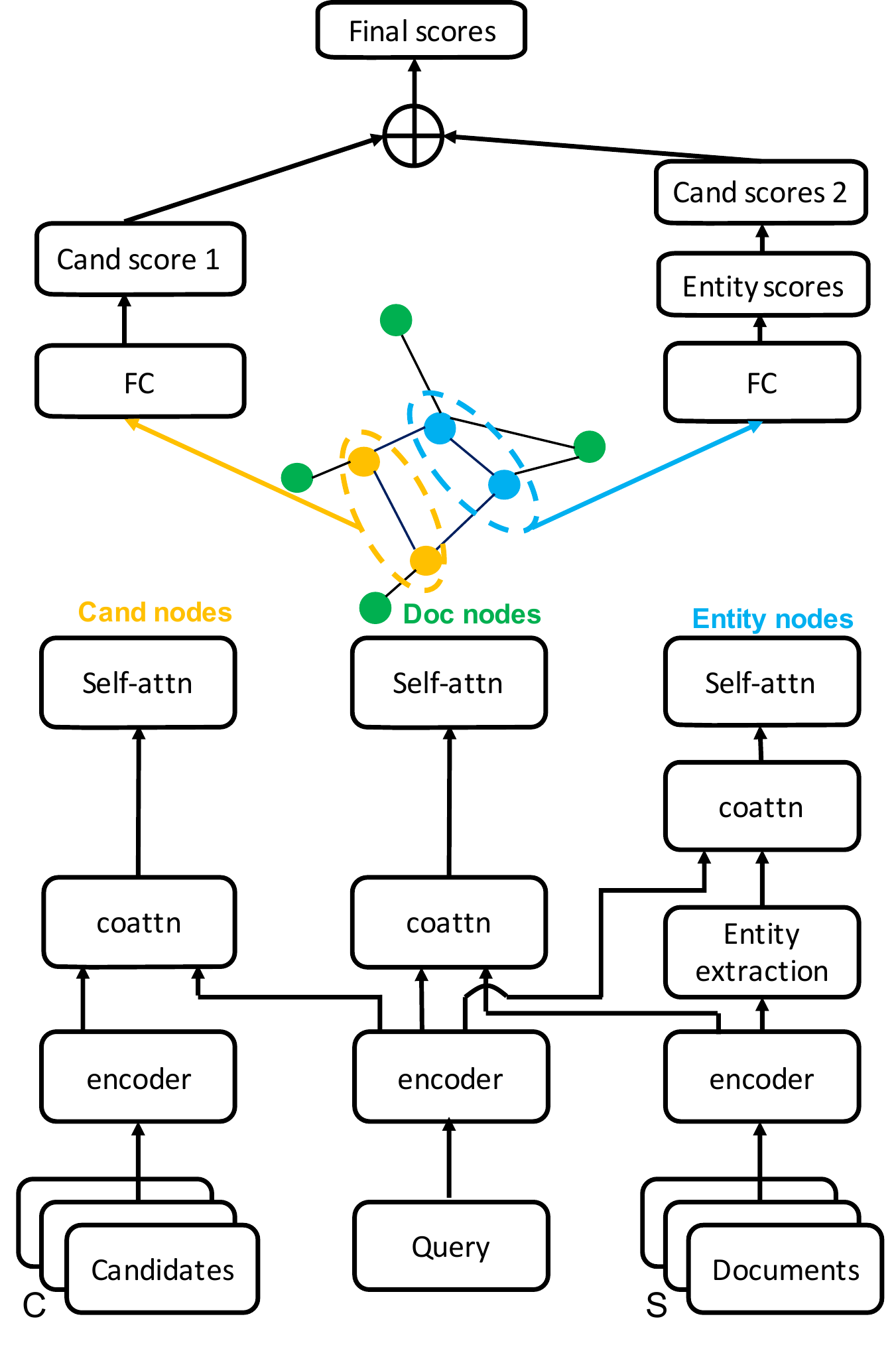}
    \caption{System diagram. $S$ and $C$ are the number of support documents and candidates respectively. We use yellow nodes to represent query-aware candidate representation, blue nodes to represent extracted query-aware entity representation and green nodes to represent query-aware document representation.}
    \label{fig:1}
\end{figure}

\textbf{Entity extraction:} entities play an import role in bridging multiple documents and connecting a query and the corresponding answer as shown in figure \ref{fig:wikihop}. For example, the entity ``get ready'' in query and two entities ``Mase'' and ``Sean Combs'' co-occur in the 2nd support document, and both ``Mase'' and ``Sean Combs'' can lead to the correct answer ``bad boy records''. Based on this observation, we propose to extract mentions of both query subject \textit{s} and candidates $C_q$ from documents. We will show later that by including mentions of query subject the performance can be improved. We use simple exact match strategy \cite{de2018question, zhong2019coarse} to find the locations of mentions of query subject and candidates, i.e. we need the start and end positions of each mention. Each mention is treated as an entity. Then, representations of entities can be taken out from the $i$-th document encoding $\mathbf{H}_s^i$. We denote an entity's representation as $\mathbf{M} \in \mathbb{R}^{l_m \times h}$ where $l_m$ is the length of the entity.

\textbf{Co-attention:} Co-attention has achieved great success for single document reading comprehension tasks \cite{seo2016bidirectional, xiong2016dynamic}, and recently was applied to multiple-hop reading comprehension \cite{zhong2019coarse}. Co-attention enables the model to 
combine learned query contextual information attended by document and document contextual information attended by query, with inputs of one query and one document. We follow the implementation of co-attention in \cite{zhong2019coarse}.

We use the co-attention between a query and a supporting document for illustration. Same operations can be applied to other documents, or between the query and extracted entities. Given RNN-encoded sequences of the query $\mathbf{H}_q \in \mathbb{R}^{l_q \times h}$  and a document $\mathbf{H}_s^i \in \mathbb{R}^{l_s^i \times h}$, the affinity matrix between the query and document can be calculated as
\begin{equation}
    \mathbf{A}_{qs}^i = \mathbf{H}_s^i (\mathbf{H}_q)^{\intercal} \in \mathbb{R}^{l_s^i \times l_q},
\end{equation}
where $\intercal$ denotes matrix transpose. Each entry of the matrix $A_{qs}^i$ indicates how related two words are, one from the query and one from the document. For simplification, in later context, we ignore the superscript $i$ which indicates the operation on the $i$-th document.

Next we derive the attention context of the query and document as follows:
\begin{equation}
    \mathbf{C}_{q} = softmax(\mathbf{A}_{qs} ^ {\intercal})\mathbf{H}_s \in \mathbb{R}^{l_q \times h},
\end{equation}
\begin{equation}
    \mathbf{C}_{s} = softmax(\mathbf{A}_{qs})\mathbf{H}_q \in \mathbb{R}^{l_s \times h}.
\end{equation}

$softmax(\cdot)$ denotes column-wise normalization. We further encode the co-attended document context using a bidirectional RNN $f$ with GRU:
\begin{equation}
    \mathbf{D}_s = f(softmax(\mathbf{A}_{qs}) C_{q}) \in \mathbb{R}^{l_s \times h}.
\end{equation}

The final co-attention context is the column-wise concatenation of $\mathbf{C}_{s}$ and $\mathbf{D}_s$: 
\begin{equation}
    \mathbf{S}_{ca} = [\mathbf{C}_{s};  \mathbf{D}_s] \in \mathbb{R}^{l_s \times 2h}.
\end{equation}

We expect $\mathbf{S}_{ca}$ carries query-aware contextual information of supporting documents as shown by \citet{zhong2019coarse}. The same co-attention module can also be applied to query and candidates, and query and entities (as shown in Figure \ref{fig:1}) to get $\mathbf{C}_{ca}$ and $\mathbf{E}_{ca}$. Note that we do not do co-attention between query and entities corresponding to query subject because query subject is already a part of the query. To keep the dimensionality consistent, we apply a single-layer multi-layer perceptron (MLP) with $tanh$ activation function to increase the dimension of the query subject entities to $2h$.

\textbf{Self-attentive pooling:} while co-attention yields a query-aware contextual representation of documents, self-attentive pooling is designed to convert the sequential contextual representation to a fixed dimensional non-sequential feature vector by selecting important query-aware information \cite{zhong2019coarse}. Self-attentive pooling summarizes the information presented in the co-attention output by calculating a score for each word in the sequence. The scores are normalized and a weighted sum based pooling is applied to the sequence to get a single feature vector as the summarization of the input sequence. Formally, the self-attention module can be formulated as the following operations given $\mathbf{S}_{ca}$ as input:
\begin{equation}
    \mathbf{a}_s = softmax(MLP(\mathbf{S}_{ca})) \in \mathbb{R}^{l_s \times 1},
\end{equation}
\begin{equation}
    \mathbf{s}_{sa} = \mathbf{a}_s ^ {\intercal} \mathbf{S}_{ca} \in \mathbb{R}^{1 \times 2h},
    \label{eq7}
\end{equation}
where $MLP(\cdot)$ is a two-layer MLP with $tanh$ as activation function. Similarly, after self-attentive pooling, we can get $\mathbf{c}_{sa}$ and $\mathbf{e}_{sa}$ for each candidate and entity.

Our context encoding module is different from the one used in \citet{zhong2019coarse} in following aspects: 1) we compute the co-attention between query and candidates which is not presented in the CFC model. 2) For entity word sequences, we first calculate co-attention with query and then use self-attention to summarize each entity word sequence while \citet{zhong2019coarse} first do self-attention on entity word sequences to get a sequence of entity vectors in each documents. Then, they apply co-attention with query.

\subsection{Reasoning over HDE graph}

\textbf{Graph building:} let a HDE graph be denoted as $\mathcal{G} = \{\mathcal{V}, \mathcal{E}\}$, where $\mathcal{V}$ stands for node representations and $\mathcal{E}$ represents edges between nodes. In our proposed HDE graph based model, we treat each document, candidate and entity extracted from documents as nodes in the HDE graph, i.e., each document (candidate/entity) corresponds to one node in the HDE graph.
These nodes represent different granularity levels of query-aware information: document nodes encode document-level global information regarding to the query; candidate nodes encode query-aware information in candidates; entity nodes encode query-aware information in specific document context or the query subject. The HDE graph is built to enable graph-based reasoning. It exploits useful structural information among query, support documents and candidates. We expect our HDE graph could perform multi-hop reasoning to locate the answer nodes or entity nodes of answers given a query.

\begin{figure}
    \centering
    \includegraphics[width=0.6\linewidth]{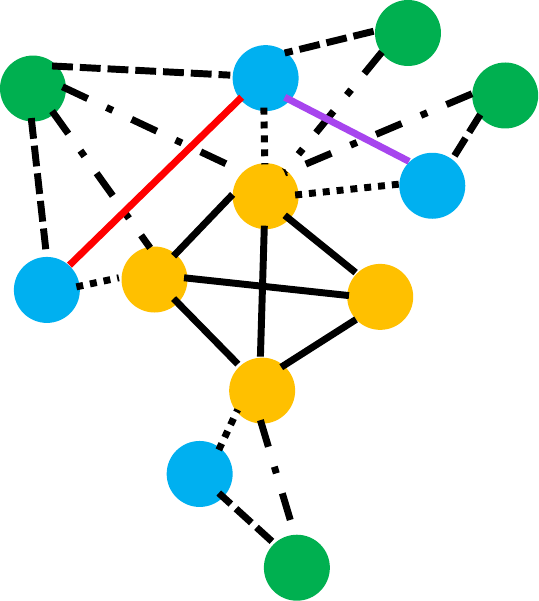}
    \caption{A toy example of HDE graph. The dash dot lines connecting documents (green nodes) and candidates (yellow nodes) correspond to type 1 edge. The normal dash lines connecting documents and entities (blue nodes) correspond to type 2 edge. The square dot lines connecting entities and candidates correspond to type 3 edge. The red solid line connecting two entities correspond to type 4 edge. The purple solid line correspond to type 5 edge. The black solid lines connecting two candidates correspond to type 6 edge. For good visualization, we ignore the type 7 edge in this figure.}
    \label{fig:2}
\end{figure}

Self-attentive pooling generates vector representations for each candidate, document and entity, which can be directly employed to initialize the node representations $\mathcal{V}$. For edge connections $\mathcal{E}$, we define the following types of edges between pairs of nodes to encode various structural information in the HDE graph:
\begin{enumerate}
    \item an edge between a document node and a candidate node if the candidate appear in the document at least one time.
    \item an edge between a document node and an entity node if the entity is extracted from the document.
    \item an edge between a candidate node and an entity node if the entity is a mention of the candidate.
    \item an edge between two entity nodes if they are extracted from the same document.
    \item an edge between two entity nodes if they are mentions of the same candidate or query subject and they are extracted from different documents.
    \item all candidate nodes connect with each other.
    \item entity nodes that do not meet previous conditions are connected. 
\end{enumerate}
Type 4, 5, 7 edges are also employed in \cite{de2018question} where the authors show the effectiveness of those different types of edges. Similarly, we treat these different edges differently to make information propagate differently over these seven different types of edges. More details will be introduced in next paragraph about message passing over the HDE graph. In 
Figure~\ref{fig:2}, we illustrate a toy example of the proposed HDE graph.

\textbf{Message passing:} we define how information propagates over the graph in order to do reasoning over the HDE graph. Different variants of GNN have different implementations of message passing strategies. In this study, we follow the message passing design in GCN \cite{kipf2016semi,de2018question} as it gives good performance on validation set compared to other strategies \cite{velivckovic2017graph, xu2018powerful}. Generally, the message passing over graphs can be achieved in two steps: aggregation and combination \cite{hamilton2017inductive}, and this process can be conducted multiple times (usually referred as layers or hops in GNN literature). Here, we give the aggregation and combination formulation of the message passing over the proposed HDE graph. The first step aggregates information from neighbors of each node, which can be formulated as

\begin{equation}
    \mathbf{z}_i ^ k = \sum_{r \in \mathcal{R}} \frac{1}{\left | \mathcal{N}_i ^ r \right |} \sum_{j \in \mathcal{N}_i ^ r} f_r(\mathbf{h}_j ^ k),
\end{equation}
where $\mathcal{R}$ is the set of all edge types, $\mathcal{N}_i ^ r$ is the neighbors of node $i$ with edge type $r$ and $\mathbf{h}_j ^ k$ is the node representation of node $j$ in layer $k$ ($\mathbf{h}_j ^ 0$ initialized with self-attention outputs). $\left | \cdot \right|$ indicates the size of the neighboring set. $f_r$ defines a transformation on the neighboring node representations, and can be implemented with a MLP. $\mathbf{z}_i ^ k$ represents the aggregated information in layer $k$ for node $i$, and can be combined with the transformed node $i$ representation:
\begin{equation}
    \mathbf{u}_i ^ k = f_s(\mathbf{h}_i ^ k) + \mathbf{z}_i ^ k,
    \label{eq8}
\end{equation}
where $f_s$ can also be implemented with a MLP.

It has been shown that GNN suffers from the smoothing problem if the number of layers is large \cite{kipf2016semi}. The smoothing problem can result in similar nodes representation and lose the discriminative ability when doing classification on nodes. To tackle this problem, we add a gating mechanism \cite{gilmer2017neural} on the combined information $\mathbf{u}_i ^ k$.
\begin{equation}
    \mathbf{g}_i ^ k = sigmoid(f_g([\mathbf{u}_i ^ k; \mathbf{h}_i ^ k]))
\end{equation}
\begin{equation}
    \mathbf{h}_i^{k+1} = tanh(\mathbf{u}_i ^ k) \odot \mathbf{g}_i ^ k + \mathbf{h}_i ^ k \odot (1-\mathbf{g}_i ^ k)
\end{equation}
$sigmoid(\cdot)$ denotes the sigmoid function on transformed concatenation of $\mathbf{u}_i ^ k$ and $\mathbf{h}_i ^ k$. $\mathbf{g}_i ^ k$ is then applied to the combined information to control the amount information from computed update or from the original node representation. $tanh(\cdot)$ functions as a non-linear activation function. $\odot$ denotes element-wise multiplication.

In this study, $f_r$, $f_s$ and $f_g$ are all implemented with single-layer MLPs, the output dimension of which is $2h$. After $K$ times message passing, all candidate, document and entity nodes will have their final updated node representation.

\subsection{Score accumulation}

The final node representations of candidate and entity nodes corresponding to mentions of candidates are used to calculate classification scores. This procedure can be formulated as
\begin{equation}
    \mathbf{a} = f_C(\mathbf{H}^C) + ACC_{max}(f_E(\mathbf{H}^E)),
    \label{eq12}
\end{equation}
where $\mathbf{H}^C \in \mathbb{R}^{C \times 2h}$ is the node representation of all candidate nodes and $C$ is the number of candidates. $\mathbf{H}^E \in \mathbb{R}^{M \times 2h}$ is the node representation of all entity nodes that correspond to candidates, and $M$ is the number of those nodes. $ACC_{max}$ is an operation that takes the maximum over scores of entities that belong to the same candidate. $f_C$ and $f_E$ are implemented with two-layer MLPs with $tanh$ activation function. The hidden layer size is half of the input dimension, and the output dimension is 1. We directly sum the scores from candidate nodes and entity nodes as the final scores over multiple candidates. Thus, the output score vector $\mathbf{a} \in \mathbb{R}^{C \times 1}$ gives a distribution over all candidates. Since the task is multi-class classification, we use cross-entropy loss as training objective which takes $\mathbf{a}$ and the labels as input.

\section{Experiments}

\subsection{Dataset}
We use \textsc{WikiHop} \cite{welbl2018constructing} to validate the effectiveness of our proposed model. The query of \textsc{WikiHop} is constructed with entities and relations from \textsc{WikiData}, while supporting documents are from \textsc{WikiReading} \cite{hewlett2016wikireading}. A bipartite graph connecting entities and documents is first built and the answer for each query is located by traversal on this graph. Candidates that are type-consistent with the answer and share the same relation in query with the answer are included, resulting in a set of candidates. Thus, \textsc{WikiHop} is a multi-choice style reading comprehension data set. There are totally about 43K samples in training set, 5K samples in development set and 2.5K samples in test set. The test set is not provided and can only be evaluated on blindly. The task is to predict the correct answer given a query and multiple supporting documents. In the experiment, we train our proposed model on all training samples in \textsc{WikiHop}, and tune model hyperparameters on all samples in development set. We only evaluate our proposed model on the unmasked version of \textsc{WikiHop}.

\subsection{Experimental settings}
Queries, support documents and candidates are tokenized into word sequences with NLTK \cite{loper2002nltk}. We empirically split the query into relation and subject entity. Exact matching strategy is employed to locate mentions of both subject entity and candidates in supporting documents. 300-dimensional GLoVe embeddings (with 840B tokens and 2.2M vocabulary size) \cite{pennington2014glove} and 100-dimensional character n-gram embeddings \cite{hashimoto2017joint} are used to convert words into 400-dimensional vector representations. Out of vocabulary words are initialized with random vectors. The embedding matrices are not updated during training. The proposed model is implemented with PyTorch \cite{paszke2017automatic}. More details about experimental and hyperparameter settings can be found in supplementary materials. The performance on development set is measured after each training epoch, and the model with the highest accuracy is saved and submitted to be evaluated on the blind test set. We will make our code publicly available after the review process.

We also prepared an ensemble model consisting of 15 models with different hyperparameter settings and random seeds. We used the simple majority voting strategy to fuse the candidate predictions of different models together.

\begin{table}[t!]
\centering
\resizebox{\linewidth}{!}{%
\begin{tabular}{l|c|c}
\hline \multirow{2}{*}{\textbf{\em Single models}} & \multicolumn{2}{c}{\textbf{Accuracy (\%)}} \\ \cline{2-3} 
 & \textbf{Dev} & \textbf{Test} \\ \hline
\hline
BiDAF & - & 42.9 \\
Coref-GRU\cite{dhingra2018neural} & 56.0 & 59.3 \\
MHQA-GRN\cite{song2018exploring} & 62.8 & 65.4 \\
Entity-GCN\cite{de2018question} & 64.8 & 67.6 \\
CFC\cite{zhong2019coarse} & 66.4 & 70.6 \\
\citet{kundu2018exploiting} & 67.1 & - \\
DynSAN\textsuperscript{*} & - & \textbf{71.4} \\
\hline
\hline
\textbf{Proposed} & \textbf{68.1} & 70.9 \\
\Xhline{3\arrayrulewidth}
\hline
\hline
\Xhline{3\arrayrulewidth} 
\textbf{\em Ensemble models} &  &  \\ \hline
\hline
Entity-GCN\cite{de2018question} & 68.5 & 71.2 \\
DynSAN\textsuperscript{*} & - & 73.8 \\
\hline
\hline
\textbf{Proposed} & \textbf{70.9} & \textbf{74.3} \\
\hline
\end{tabular}}
\caption{Performance comparison among different models on \textsc{WikiHop} development and test set. The results of ``BiDAF'' are presented in the paper by \citet{welbl2018constructing}. Models annotated with ``*'' are unpublished but available on \textsc{WikiHop} leaderboard. ``-'' indicates unavailable numbers.}
\label{table:1}
\end{table}

\begin{table}[t]
\centering
\begin{tabular}{l|c|c}
\hline
\multirow{2}{*}{\textbf{Model}} & \multicolumn{2}{c}{\textbf{Accuracy (\%)}} \\ \cline{2-3} 
 & \textbf{Dev} & $\boldsymbol\Delta$ \\ \hline \hline
Full model & \textbf{68.1} & - \\ \hline
\hline
~~- HDE graph & 65.5 & 2.6 \\ \hline
~~- different edge types & 66.7 & 1.4 \\ \hline
~~- candidate nodes scores & 67.1 & 1.0 \\ \hline
~~- entity nodes scores & 66.6 & 1.5 \\ \hline
~~- candidate nodes & 66.2 & 1.9 \\ \hline
~~- document nodes & 67.6 & 0.5 \\ \hline
~~- entity nodes & 63.6 & 4.5 \\ \hline
\end{tabular}
\caption{Ablation results on the \textsc{WikiHop} dev set.}
\label{table:2}
\end{table}

\subsection{Results}
In Table~\ref{table:1}, we show the results of the our proposed HDE graph based model on both development and test set and compare it with previously published results. We show that our proposed HDE graph based model improves the published state-of-the-art accuracy on development set from 67.1\% \cite{kundu2018exploiting} to 68.1\%, on the blind test set from 70.6\% \cite{zhong2019coarse} to 70.9\%. Compared to the best single model ``DynSAN'' (unpublished) on \textsc{WikiHop} leaderboard, our proposed model is still 0.5\% worse. Compared to two previous studies using GNN for multi-hop reading comprehension \cite{song2018exploring, de2018question}, our model surpasses them by a large margin even though we do not use better pre-trained contextual embedding ELMo \cite{peters2018deep}.

For the ensemble models, our proposed system achieves the state-of-the-art performance, which is also 0.2\% higher than the reported human performance \cite{welbl2018constructing}. Even though our single model is a little worse than the ``DynSAN'', our ensemble model is better than both the ensembled ``DynSAN'' and the ensembled ``Entity-GCN''.




\begin{table}[t]
\resizebox{\linewidth}{!}{%
\begin{tabular}{|c|c|c|}
\hline
Model & Single-follow & Multi-follow \\ \hline
With HDE graph & 67.8 & 71.0 \\ \hline
Without HDE graph & 66.7 & 67.0 \\ \hline
\end{tabular}}
\caption{Accuracy(\%) comparison under different types of samples.}
\label{table:3}
\end{table}

\subsection{Ablation studies} 
In order to better understand the contribution of different modules to the performance, we conduct several ablation studies on the development set of \textsc{WikiHop}.

If we remove the proposed HDE graph and directly use the representations of candidates and entities corresponding to mentions of candidates (equation \ref{eq7}) for score accumulation, the accuracy on \textsc{WikiHop} development set drops 2.6\% absolutely. This proves the efficacy of the proposed HDE graph on multi-hop reasoning across multiple documents.

If we treat all edge types equally without using different GNN parameters for different edge types (equation \ref{eq8}), the accuracy drops 1.4\%, which indicates that different information encoded by different types of edges is also important to retain good performance;
If only scores of entity nodes (right part of equation \ref{eq12}) are considered in score accumulation, the accuracy on dev set degrades by 1.0\%; if only scores of candidates nodes (left part of equation \ref{eq12}) are considered, the accuracy degrades by 1.5\%. This means that the scores on entity nodes contribute more to the classification, which is reasonable because entities carry context information in the document while candidates do not.

We also investigate the effect of removing different types of nodes. Note that removing nodes is not the same as removing scores from candidate/entity nodes --- it means we do not use the scores on these nodes during score accumulation but nodes still exist during message passing on the HDE graph. However, removing one type of nodes means the nodes and corresponding edges do not exist in the HDE graph. The ablation shows that removing entity nodes results in the largest degradation of performance while removing document nodes result in the least degradation. This finding is consistent with the study by \cite{de2018question} where they emphasize the importance of entities in multi-hop reasoning. The small contribution of document nodes is probably caused by too much information loss during self-attentive pooling over long sequences.  Better ways are needed to encode document information into graph. More ablation studies are included in the supplementary materials due to space constraint.

\begin{figure}[t]
  \centering
  \begin{minipage}{\linewidth}
    \includegraphics[width=\linewidth]{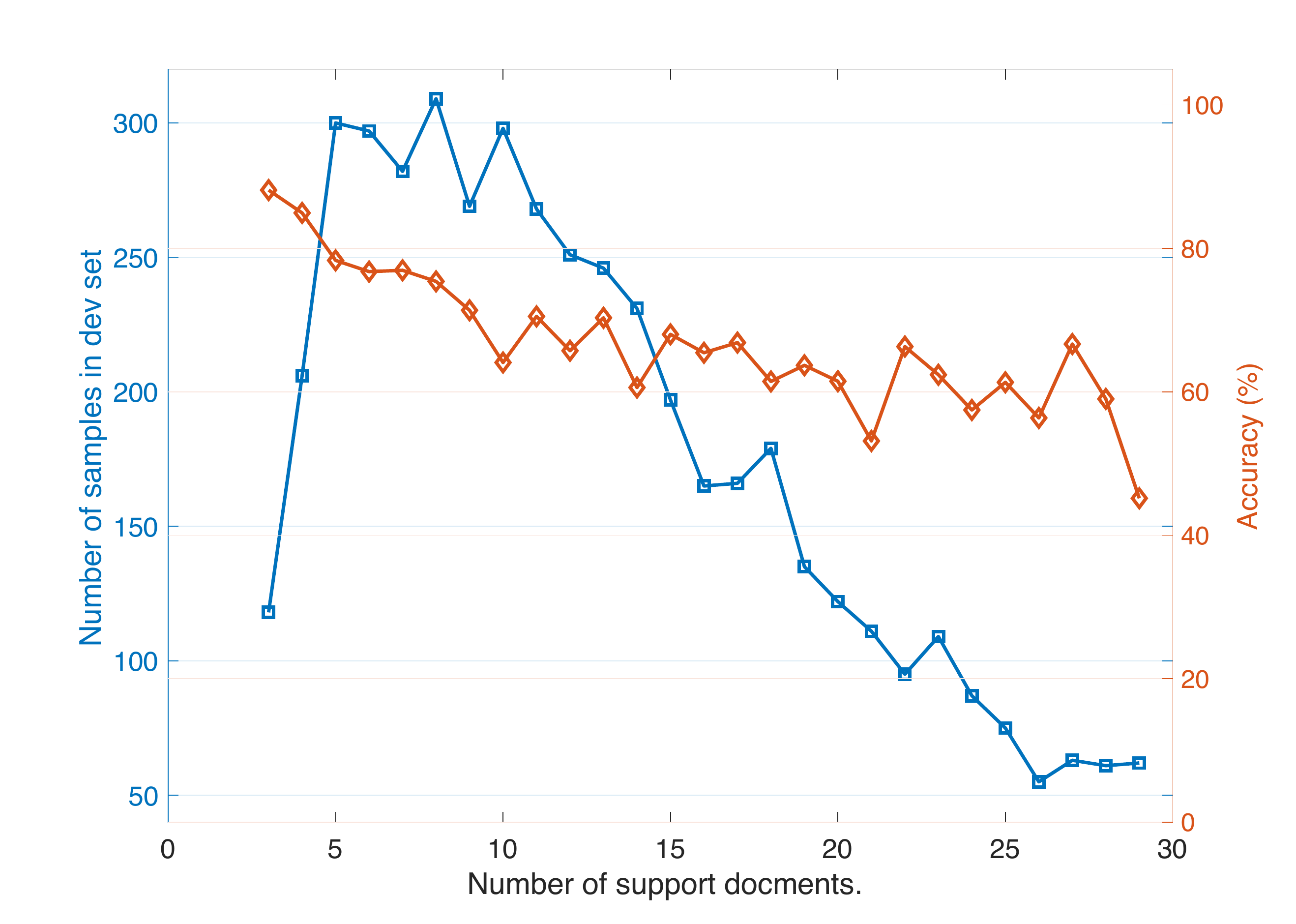}
    \caption{Plots between number of support documents (x-axis) and number of examples (left y-axis), and between number of support documents and accuracy (right y-axis).}
    \label{fig:5}
    \end{minipage}%
    \newline
    \begin{minipage}{\linewidth}
    \includegraphics[width=\linewidth]{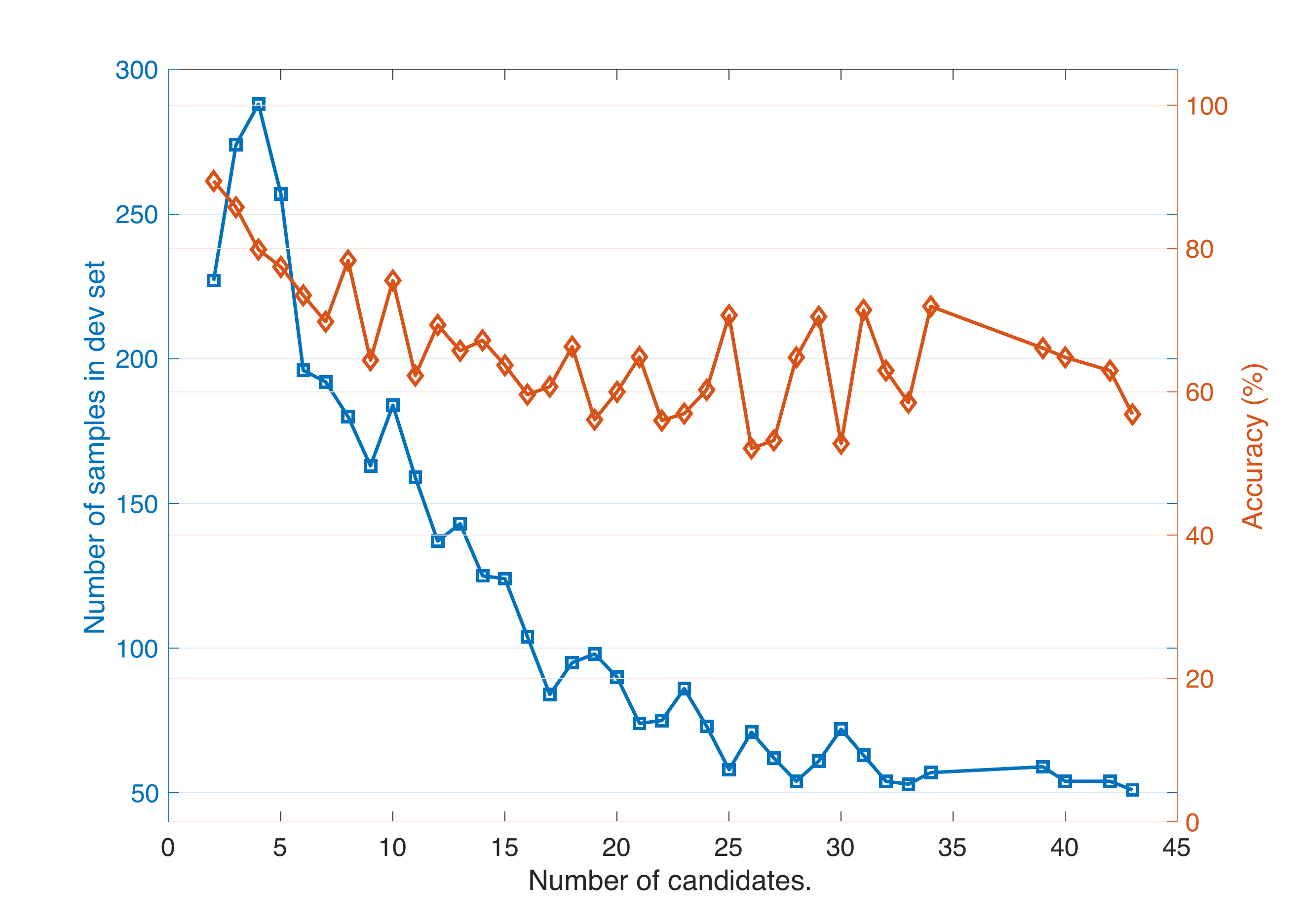}
    \caption{Plots between number of candidates (x-axis) and number of examples (left y-axis), and between number of candidates and accuracy (right y-axis).}
    \label{fig:6}
\end{minipage} 
\end{figure}

\subsection{Result analysis}
\label{sec:res_analysis}


To investigate how the HDE graph helps multi-hop reasoning, we conduct experiments on \textsc{WikiHop} development set where we discard the HDE graph and only use the candidate and entity representations output by self-attention. In Table \ref{table:3}, ``Single-follow'' (2069 samples in the dev set) means a single document is enough to answer the query, while ``Multi-follow'' (2601 samples) means multiple documents are needed. These information is provided in \cite{welbl2018constructing}. We observe in Table \ref{table:2} that the performance is consistently better for ``with HDE graph'' in both cases. In  ``Single-follow'' case the absolute accuracy improvement is 1.1\%, while a significant 4.0\% improvement is achieved in the ``Multi-follow'' case, which has even more samples than ``Single-follow'' case. This proves that the proposed HDE graph is good at reasoning over multiple documents. 



We also investigate how our model performs w.r.t. the number of support documents and number of candidates given an input sample. In Figure~\ref{fig:5}, the blue line with square markers shows the number of support documents in one sample (x-axis) and the corresponding frequencies in the development set (y-axis). The orange line with diamond markers shows the change of accuracy with the increasing of number of support documents. We choose the number of support documents with more than 50 appearances in the development set. For example, there are about 300 samples with 5 support documents and the accuracy of our model on these 300 samples is about 80\%. Overall, we find the accuracy decreases with the increasing number of support documents. 
This is reasonable because more documents possibly means more entities and bigger graph, and is more challenging for reasoning. Figure~\ref{fig:6} indicates the similar trend (when the number of candidates are less than 20) with the increasing number of candidates, which we believe is partly caused by the larger HDE graph. Also, more candidates cause more confusion in the selection.

\section{Conclusion}
We propose a new GNN-based method for multi-hop RC across multiple documents. We introduce the HDE graph, a heterogeneous graph for multiple-hop reasoning over nodes representing different granularity levels of information. We use co-attention and self-attention to encode candidates, documents, entities of mentions of candidates and query subjects into query-aware representations, which are then employed to initialize graph node representations. Evaluated on \textsc{WikiHop}, our end-to-end trained single neural model delivers competitive results while our ensemble model achieves the state-of-the-art performance. In the future, we would like to investigate explainable GNN for this task, such as explicit reasoning path in \cite{kundu2018exploiting}, and work on other data sets such as HotpotQA.


\section{Acknowledgements}
We would like to thank Johannes Welbl from University College London for running evaluation on our submitted model.

\bibliography{references}

\begin{thebibliography}{37}
\expandafter\ifx\csname natexlab\endcsname\relax\def\natexlab#1{#1}\fi

\bibitem[{Bastings et~al.(2017)Bastings, Titov, Aziz, Marcheggiani, and
  Simaan}]{bastings2017graph}
Joost Bastings, Ivan Titov, Wilker Aziz, Diego Marcheggiani, and Khalil Simaan.
  2017.
\newblock Graph convolutional encoders for syntax-aware neural machine
  translation.
\newblock In \emph{Proceedings of the 2017 Conference on Empirical Methods in
  Natural Language Processing}, pages 1957--1967.

\bibitem[{Cho et~al.(2014)Cho, van Merrienboer, Gulcehre, Bougares, Schwenk,
  and Bengio}]{cho2014learning}
Kyunghyun Cho, B~van Merrienboer, Caglar Gulcehre, F~Bougares, H~Schwenk, and
  Yoshua Bengio. 2014.
\newblock Learning phrase representations using rnn encoder-decoder for
  statistical machine translation.
\newblock In \emph{Conference on Empirical Methods in Natural Language
  Processing (EMNLP 2014)}.

\bibitem[{De~Cao et~al.(2018)De~Cao, Aziz, and Titov}]{de2018question}
Nicola De~Cao, Wilker Aziz, and Ivan Titov. 2018.
\newblock Question answering by reasoning across documents with graph
  convolutional networks.
\newblock \emph{arXiv preprint arXiv:1808.09920}.

\bibitem[{Devlin et~al.(2018)Devlin, Chang, Lee, and
  Toutanova}]{devlin2018bert}
Jacob Devlin, Ming-Wei Chang, Kenton Lee, and Kristina Toutanova. 2018.
\newblock Bert: Pre-training of deep bidirectional transformers for language
  understanding.
\newblock \emph{arXiv preprint arXiv:1810.04805}.

\bibitem[{Dhingra et~al.(2018)Dhingra, Jin, Yang, Cohen, and
  Salakhutdinov}]{dhingra2018neural}
Bhuwan Dhingra, Qiao Jin, Zhilin Yang, William Cohen, and Ruslan Salakhutdinov.
  2018.
\newblock Neural models for reasoning over multiple mentions using coreference.
\newblock In \emph{Proceedings of the 2018 Conference of the North American
  Chapter of the Association for Computational Linguistics: Human Language
  Technologies, Volume 2 (Short Papers)}, volume~2, pages 42--48.

\bibitem[{Gilmer et~al.(2017)Gilmer, Schoenholz, Riley, Vinyals, and
  Dahl}]{gilmer2017neural}
Justin Gilmer, Samuel~S Schoenholz, Patrick~F Riley, Oriol Vinyals, and
  George~E Dahl. 2017.
\newblock Neural message passing for quantum chemistry.
\newblock In \emph{Proceedings of the 34th International Conference on Machine
  Learning-Volume 70}, pages 1263--1272. JMLR. org.

\bibitem[{Hamilton et~al.(2017)Hamilton, Ying, and
  Leskovec}]{hamilton2017inductive}
Will Hamilton, Zhitao Ying, and Jure Leskovec. 2017.
\newblock Inductive representation learning on large graphs.
\newblock In \emph{Advances in Neural Information Processing Systems}, pages
  1024--1034.

\bibitem[{Hashimoto et~al.(2017)Hashimoto, Tsuruoka, Socher
  et~al.}]{hashimoto2017joint}
Kazuma Hashimoto, Yoshimasa Tsuruoka, Richard Socher, et~al. 2017.
\newblock A joint many-task model: Growing a neural network for multiple nlp
  tasks.
\newblock In \emph{Proceedings of the 2017 Conference on Empirical Methods in
  Natural Language Processing}, pages 1923--1933.

\bibitem[{Hewlett et~al.(2016)Hewlett, Lacoste, Jones, Polosukhin, Fandrianto,
  Han, Kelcey, and Berthelot}]{hewlett2016wikireading}
Daniel Hewlett, Alexandre Lacoste, Llion Jones, Illia Polosukhin, Andrew
  Fandrianto, Jay Han, Matthew Kelcey, and David Berthelot. 2016.
\newblock Wikireading: A novel large-scale language understanding task over
  wikipedia.
\newblock \emph{arXiv preprint arXiv:1608.03542}.

\bibitem[{Khashabi et~al.(2018)Khashabi, Chaturvedi, Roth, Upadhyay, and
  Roth}]{MultiRC2018}
Daniel Khashabi, Snigdha Chaturvedi, Michael Roth, Shyam Upadhyay, and Dan
  Roth. 2018.
\newblock Looking beyond the surface:a challenge set for reading comprehension
  over multiple sentences.
\newblock In \emph{Proceedings of North American Chapter of the Association for
  Computational Linguistics (NAACL)}.

\bibitem[{Kipf and Welling(2016)}]{kipf2016semi}
Thomas~N Kipf and Max Welling. 2016.
\newblock Semi-supervised classification with graph convolutional networks.
\newblock \emph{arXiv preprint arXiv:1609.02907}.

\bibitem[{Ko{\v{c}}isk{\`y} et~al.(2018)Ko{\v{c}}isk{\`y}, Schwarz, Blunsom,
  Dyer, Hermann, Melis, and Grefenstette}]{kovcisky2018narrativeqa}
Tom{\'a}{\v{s}} Ko{\v{c}}isk{\`y}, Jonathan Schwarz, Phil Blunsom, Chris Dyer,
  Karl~Moritz Hermann, G{\'a}abor Melis, and Edward Grefenstette. 2018.
\newblock The narrativeqa reading comprehension challenge.
\newblock \emph{Transactions of the Association of Computational Linguistics},
  6:317--328.

\bibitem[{Kundu et~al.(2018)Kundu, Khot, and Sabharwal}]{kundu2018exploiting}
Souvik Kundu, Tushar Khot, and Ashish Sabharwal. 2018.
\newblock Exploiting explicit paths for multi-hop reading comprehension.
\newblock \emph{arXiv preprint arXiv:1811.01127}.

\bibitem[{Liu et~al.(2018)Liu, Chang, Huang, Tang, and
  Cheung}]{liu2018contextualized}
Pengfei Liu, Shuaichen Chang, Xuanjing Huang, Jian Tang, and Jackie Chi~Kit
  Cheung. 2018.
\newblock Contextualized non-local neural networks for sequence learning.
\newblock \emph{arXiv preprint arXiv:1811.08600}.

\bibitem[{Loper and Bird(2002)}]{loper2002nltk}
Edward Loper and Steven Bird. 2002.
\newblock Nltk: the natural language toolkit.
\newblock \emph{arXiv preprint cs/0205028}.

\bibitem[{Marcheggiani et~al.(2018)Marcheggiani, Bastings, and
  Titov}]{marcheggiani2018exploiting}
Diego Marcheggiani, Joost Bastings, and Ivan Titov. 2018.
\newblock Exploiting semantics in neural machine translation with graph
  convolutional networks.
\newblock In \emph{Proceedings of the 2018 Conference of the North American
  Chapter of the Association for Computational Linguistics: Human Language
  Technologies, Volume 2 (Short Papers)}, volume~2, pages 486--492.

\bibitem[{Mihaylov et~al.(2018)Mihaylov, Clark, Khot, and
  Sabharwal}]{mihaylov2018can}
Todor Mihaylov, Peter Clark, Tushar Khot, and Ashish Sabharwal. 2018.
\newblock Can a suit of armor conduct electricity? a new dataset for open book
  question answering.
\newblock \emph{arXiv preprint arXiv:1809.02789}.

\bibitem[{Paszke et~al.(2017)Paszke, Gross, Chintala, Chanan, Yang, DeVito,
  Lin, Desmaison, Antiga, and Lerer}]{paszke2017automatic}
Adam Paszke, Sam Gross, Soumith Chintala, Gregory Chanan, Edward Yang, Zachary
  DeVito, Zeming Lin, Alban Desmaison, Luca Antiga, and Adam Lerer. 2017.
\newblock Automatic differentiation in pytorch.

\bibitem[{Pennington et~al.(2014)Pennington, Socher, and
  Manning}]{pennington2014glove}
Jeffrey Pennington, Richard Socher, and Christopher Manning. 2014.
\newblock Glove: Global vectors for word representation.
\newblock In \emph{Proceedings of the 2014 conference on empirical methods in
  natural language processing (EMNLP)}, pages 1532--1543.

\bibitem[{Peters et~al.(2018)Peters, Neumann, Iyyer, Gardner, Clark, Lee, and
  Zettlemoyer}]{peters2018deep}
Matthew Peters, Mark Neumann, Mohit Iyyer, Matt Gardner, Christopher Clark,
  Kenton Lee, and Luke Zettlemoyer. 2018.
\newblock Deep contextualized word representations.
\newblock In \emph{Proceedings of the 2018 Conference of the North American
  Chapter of the Association for Computational Linguistics: Human Language
  Technologies, Volume 1 (Long Papers)}, volume~1, pages 2227--2237.

\bibitem[{Rajpurkar et~al.(2018)Rajpurkar, Jia, and Liang}]{rajpurkar2018know}
Pranav Rajpurkar, Robin Jia, and Percy Liang. 2018.
\newblock Know what you don’t know: Unanswerable questions for squad.
\newblock In \emph{Proceedings of the 56th Annual Meeting of the Association
  for Computational Linguistics (Volume 2: Short Papers)}, volume~2, pages
  784--789.

\bibitem[{Rajpurkar et~al.(2016)Rajpurkar, Zhang, Lopyrev, and
  Liang}]{rajpurkar2016squad}
Pranav Rajpurkar, Jian Zhang, Konstantin Lopyrev, and Percy Liang. 2016.
\newblock Squad: 100,000+ questions for machine comprehension of text.
\newblock In \emph{Proceedings of the 2016 Conference on Empirical Methods in
  Natural Language Processing}, pages 2383--2392.

\bibitem[{Reddy et~al.(2018)Reddy, Chen, and Manning}]{reddy2018coqa}
Siva Reddy, Danqi Chen, and Christopher~D Manning. 2018.
\newblock Coqa: A conversational question answering challenge.
\newblock \emph{arXiv preprint arXiv:1808.07042}.

\bibitem[{Seo et~al.(2016)Seo, Kembhavi, Farhadi, and
  Hajishirzi}]{seo2016bidirectional}
Minjoon Seo, Aniruddha Kembhavi, Ali Farhadi, and Hannaneh Hajishirzi. 2016.
\newblock Bidirectional attention flow for machine comprehension.
\newblock \emph{arXiv preprint arXiv:1611.01603}.

\bibitem[{Song et~al.(2018)Song, Wang, Yu, Zhang, Florian, and
  Gildea}]{song2018exploring}
Linfeng Song, Zhiguo Wang, Mo~Yu, Yue Zhang, Radu Florian, and Daniel Gildea.
  2018.
\newblock Exploring graph-structured passage representation for multi-hop
  reading comprehension with graph neural networks.
\newblock \emph{arXiv preprint arXiv:1809.02040}.

\bibitem[{Sun et~al.(2018)Sun, Yu, Yu, and Cardie}]{sun2018improving}
Kai Sun, Dian Yu, Dong Yu, and Claire Cardie. 2018.
\newblock Improving machine reading comprehension with general reading
  strategies.
\newblock \emph{arXiv preprint arXiv:1810.13441}.

\bibitem[{Tay et~al.(2018)Tay, Luu, Hui, and Su}]{tay2018densely}
Yi~Tay, Anh~Tuan Luu, Siu~Cheung Hui, and Jian Su. 2018.
\newblock Densely connected attention propagation for reading comprehension.
\newblock In \emph{Advances in Neural Information Processing Systems}, pages
  4911--4922.

\bibitem[{Veli{\v{c}}kovi{\'c} et~al.(2017)Veli{\v{c}}kovi{\'c}, Cucurull,
  Casanova, Romero, Lio, and Bengio}]{velivckovic2017graph}
Petar Veli{\v{c}}kovi{\'c}, Guillem Cucurull, Arantxa Casanova, Adriana Romero,
  Pietro Lio, and Yoshua Bengio. 2017.
\newblock Graph attention networks.
\newblock \emph{arXiv preprint arXiv:1710.10903}.

\bibitem[{Welbl et~al.(2018)Welbl, Stenetorp, and
  Riedel}]{welbl2018constructing}
Johannes Welbl, Pontus Stenetorp, and Sebastian Riedel. 2018.
\newblock Constructing datasets for multi-hop reading comprehension across
  documents.
\newblock \emph{Transactions of the Association of Computational Linguistics},
  6:287--302.

\bibitem[{Weston et~al.(2015)Weston, Bordes, Chopra, Rush, van Merri{\"e}nboer,
  Joulin, and Mikolov}]{weston2015towards}
Jason Weston, Antoine Bordes, Sumit Chopra, Alexander~M Rush, Bart van
  Merri{\"e}nboer, Armand Joulin, and Tomas Mikolov. 2015.
\newblock Towards ai-complete question answering: A set of prerequisite toy
  tasks.
\newblock \emph{arXiv preprint arXiv:1502.05698}.

\bibitem[{Wu et~al.(2019)Wu, Socher, and Xiong}]{wu2019global}
Chien-Sheng Wu, Richard Socher, and Caiming Xiong. 2019.
\newblock Global-to-local memory pointer networks for task-oriented dialogue.
\newblock \emph{arXiv preprint arXiv:1901.04713}.

\bibitem[{Xiong et~al.(2016)Xiong, Zhong, and Socher}]{xiong2016dynamic}
Caiming Xiong, Victor Zhong, and Richard Socher. 2016.
\newblock Dynamic coattention networks for question answering.
\newblock \emph{arXiv preprint arXiv:1611.01604}.

\bibitem[{Xu et~al.(2018)Xu, Hu, Leskovec, and Jegelka}]{xu2018powerful}
Keyulu Xu, Weihua Hu, Jure Leskovec, and Stefanie Jegelka. 2018.
\newblock How powerful are graph neural networks?
\newblock \emph{arXiv preprint arXiv:1810.00826}.

\bibitem[{Yang et~al.(2018)Yang, Qi, Zhang, Bengio, Cohen, Salakhutdinov, and
  Manning}]{yang2018hotpotqa}
Zhilin Yang, Peng Qi, Saizheng Zhang, Yoshua Bengio, William Cohen, Ruslan
  Salakhutdinov, and Christopher~D Manning. 2018.
\newblock Hotpotqa: A dataset for diverse, explainable multi-hop question
  answering.
\newblock In \emph{Proceedings of the 2018 Conference on Empirical Methods in
  Natural Language Processing}, pages 2369--2380.

\bibitem[{Yao et~al.(2018)Yao, Mao, and Luo}]{yao2018graph}
Liang Yao, Chengsheng Mao, and Yuan Luo. 2018.
\newblock Graph convolutional networks for text classification.
\newblock \emph{arXiv preprint arXiv:1809.05679}.

\bibitem[{Zhang et~al.(2018)Zhang, Qi, and Manning}]{zhang2018graph}
Yuhao Zhang, Peng Qi, and Christopher~D Manning. 2018.
\newblock Graph convolution over pruned dependency trees improves relation
  extraction.
\newblock In \emph{Proceedings of the 2018 Conference on Empirical Methods in
  Natural Language Processing}, pages 2205--2215.

\bibitem[{Zhong et~al.(2019)Zhong, Xiong, Keskar, and Socher}]{zhong2019coarse}
Victor Zhong, Caiming Xiong, Nitish~Shirish Keskar, and Richard Socher. 2019.
\newblock Coarse-grain fine-grain coattention network for multi-evidence
  question answering.
\newblock \emph{arXiv preprint arXiv:1901.00603}.

\end{thebibliography}
\bibliographystyle{acl_natbib}

\end{document}